%% file: main.tex
\pdfoutput=1
\documentclass[11pt]{article}

\usepackage[final]{acl}

\usepackage{times}
\usepackage{latexsym}
\usepackage{booktabs}
\usepackage{multirow}
\usepackage{array}
\usepackage{xcolor}
\usepackage{tikz}
\usepackage{colortbl}
\usepackage[T1]{fontenc}
\usepackage{tgheros} % Helvetica-like font that works with pdfLaTeX
\usepackage{listings}

\usepackage[T1]{fontenc}
\usepackage[utf8]{inputenc}

\usepackage{microtype}
\usepackage{amsmath}

\usepackage{inconsolata}

\usepackage{graphicx}

\title{Paying Alignment Tax  with Contrastive Learning}
\author{Buse Sibel Korkmaz \\
  Imperial College London \\ London, UK \\
  \texttt{buse.korkmaz18@imperial.ac.uk} \\\And
  Rahul Nair \\
  IBM Research Europe \\ Dublin, Ireland \\
  \texttt{rahul.nair@ie.ibm.com} \\ \AND 
  Elizabeth Daly \\ 
  IBM Research Europe \\ Dublin, Ireland \\
   \texttt{elizabeth.daly@ie.ibm.com} \\\And 
  Antonio del Rio Chanona \\ 
  Imperial College London \\ London, UK \\
   \texttt{a.del-rio-chanona@imperial.ac.uk}}

\begin{document}
\maketitle
\begin{abstract}
 Current debiasing approaches often result a degradation in model capabilities such as factual accuracy and knowledge retention. Through systematic evaluation across multiple benchmarks, we demonstrate that existing debiasing methods face fundamental trade-offs, particularly in smaller models, leading to reduced truthfulness, knowledge loss, or unintelligible outputs. To address these limitations, we propose a contrastive learning framework that learns through carefully constructed positive and negative examples. Our approach introduces contrast computation and dynamic loss scaling to balance bias mitigation with faithfulness preservation. Experimental results across multiple model scales demonstrate that our method achieves substantial improvements in both toxicity reduction and faithfulness preservation. Most importantly, we show that our framework is the first to consistently improve both metrics simultaneously, avoiding the capability degradation characteristic of existing approaches. These results suggest that explicit modeling of both positive and negative examples through contrastive learning could be a promising direction for reducing the alignment tax in language model debiasing.
\end{abstract}

\section{Introduction}

Large language models (LLMs) have demonstrated remarkable capabilities in various tasks. 
% from text generation to complex reasoning. 
However, these models can perpetuate and amplify societal biases present in their training data, leading to potentially harmful outputs. Although numerous bias mitigation techniques have been proposed \citep{ravfogel-etal-2020-null, liang-etal-2020-towards, webster2020measuring, schick-etal-2021-self, morabito-etal-2023-debiasing}, there is a fundamental tension between reducing harmful biases and maintaining model performance on other important dimensions, such as factual accuracy and general knowledge.

Recent work in bias mitigation has focused mainly on improving performance on specific bias-related benchmarks, often through methods such as data filtering, fine-tuning with debiased datasets \citep{zmigrod-etal-2019-counterfactual, dinan-etal-2020-queens}, or prompt-based approaches \citep{morabito-etal-2023-debiasing}. However, our systematic analysis in Section \ref{sec:analysis} reveals that these improvements often come at a significant cost: what is commonly termed as the \emph{``alignment tax''} \citep{lin-etal-2024-mitigating}. This phenomenon manifests as a degradation in model performance across crucial capabilities, particularly in smaller models, where the trade-off between bias mitigation and maintaining general knowledge becomes especially pronounced due to limited capability of modelling.

Our analysis demonstrates that existing debiasing methods often lead to one of three undesirable outcomes: (1) generation of unintelligible or overly cautious text, (2) complete abstention from answering potentially sensitive questions, or (3) decreased faithfulness to factual information. These issues are particularly evident when models are evaluated in multiple dimensions simultaneously, such as toxicity reduction \citep{gehman-etal-2020-realtoxicityprompts} and factual accuracy \citep{lin-etal-2022-truthfulqa}, where current approaches struggle to maintain performance in both metrics.

To substantiate these claims, we present a comprehensive evaluation framework that measures the alignment tax across multiple dimensions. Using this framework, we first demonstrate a consistent negative correlation between toxicity reduction and performance on truthfulness benchmarks, indicating a fundamental trade-off that existing approaches have not successfully resolved. Our detailed analysis on the MMLU-Pro benchmark \citep{hendrycks2021measuring} further reveals specific knowledge categories that are disproportionately affected by debiasing interventions, highlighting the non-uniform nature of the alignment tax across different domains of knowledge. Moreover, when examining practical applications such as Reddit TL;DR summarization \citep{volske-etal-2017-tl}, we find that no existing method successfully maintains faithfulness while reducing toxic content, underscoring the challenges of applications of existing debiasing methods in the real world.

% I will add GitHub link
To address these limitations, we propose a novel framework based on contrastive learning that simultaneously improves both bias mitigation and factual accuracy. Our approach builds upon recent work in contrastive learning for bias mitigation \citep{li-etal-2023-prompt, park-etal-2024-contrastive} and faithfulness \citep{cao-wang-2021-cliff}, but differs by explicitly modeling faithfulness and bias mitigation together through carefully constructed positive and negative example pairs. This is achieved through a combination of data augmentation techniques and a contrastive learning objective that helps models learn to distinguish between biased and unbiased outputs while maintaining factual accuracy. The main contributions of this work are:
\begin{itemize}
    \item A comprehensive benchmark for measuring the alignment tax in bias mitigation algorithms, enabling systematic evaluation of trade-offs between bias reduction and other important model capabilities
    % \item Novel data augmentation techniques for generating high-quality positive and negative training pairs for contrastive learning in the context of bias mitigation
    \item A contrastive learning framework including contrastive data augmentation that demonstrably improves both bias metrics and factual accuracy, along with open-source implementation
    \item Empirical evidence showing improvements across multiple metrics without the severe trade-offs characteristic of existing approaches
\end{itemize}

Our findings on the negative correlations between language modeling quality and debiasing score are consistent with prior work \citep{meade-etal-2022-empirical}. However, our results demonstrate that it is possible to achieve meaningful bias reduction without sacrificing model performance on other important tasks when using our contrastive learning approach.
% We make our data, code and used models publicly available \footnote{GitHub link will be released upon acceptance.}. 

\section{Related Work}\label{sec:relatedwork}

Prior work in bias mitigation has explored a variety of techniques targeting different aspects of model behavior. At the representation level, INLP \citep{ravfogel-etal-2020-null} operates by iteratively identifying and removing directions in the embedding space that encode protected attributes, aiming to eliminate bias while preserving other semantic information. Sentence Debiasing \citep{liang-etal-2020-towards} takes a more direct approach by modifying sentence representations themselves, identifying and neutralizing bias-carrying components while attempting to maintain semantic content and fluency. Other approaches leverage training dynamics, such as Dropout-based debiasing \citep{webster2020measuring}, which uses stochastic noise to reduce reliance on spurious correlations.

More recent methods have focused on prompt-based strategies. Self Debiasing \citep{schick-etal-2021-self} guides models to recognize and avoid biased content through carefully crafted prompts without requiring model fine-tuning. Building on this, Instructive Debiasing \citep{morabito-etal-2023-debiasing} incorporates explicit bias avoidance instructions into the training process. Counterfactual Data Augmentation \citep{zmigrod-etal-2019-counterfactual, dinan-etal-2020-queens} approaches the problem through data manipulation, creating balanced datasets by systematically varying protected attributes.

Contrastive learning is a promising direction for both faithfulness preservation and bias mitigation. In the faithfulness domain, CLIFF \citep{cao-wang-2021-cliff} demonstrated the effectiveness of contrastive learning for improving factual consistency. For bias mitigation specifically, recent work has explored contrastive prompt tuning \citep{li-etal-2023-prompt}, contrastive debiasing \citep{park-etal-2024-contrastive}, and applications to speech \citep{cl-to-mitigate-bias-in-speech}. While these works tackle faithfulness and debiasing independently, our work combines the two in a joint loss computation.
%Our work extends these approaches by introducing faithfulness and toxicity based contrast computation and dynamic loss scaling in the presence of strong negatives.

% The increasing focus on factuality in language models has led to the development of specialized detectors \citep{achintalwar2024detectors}, which inform our evaluation methodology. Our framework uniquely combines insights from these various approaches to simultaneously address bias and faithfulness, introducing novel techniques for data augmentation and contrastive learning while carefully considering the trade-offs identified in previous work.

\section{Alignment Tax in Debiasing Methods}\label{sec:analysis}

%In this section, 
We systematically evaluate different debiasing approaches introduced in Section \ref{sec:relatedwork} across multiple dimensions of model performance to demonstrate shortcomings of current debiasing techniques.

\subsection{Models}

\input{truthfulqa_vs_toxicity}

%To evaluate these debiasing methods, we conduct experiments using 
We use three representative language models spanning different sizes and architectures, specifically %We use 
\texttt{Llama2} \citep{touvron2023llama}, a 7B parameter model representing the open-source large language models; \texttt{GPT2} \citep{radford2019language}, a widely-used smaller model that serves as a baseline for comparison; and \texttt{Phi2} \citep{gunasekar2023textbooks} from the Phi model family, which offers an intermediate scale for our analysis.

For each base model, we create variants using each of the debiasing methods described above, resulting in a comprehensive test suite that allows us to analyze the impact of different bias mitigation strategies across model scales. This experimental setup enables us to examine not only how different debiasing approaches perform in isolation, but also how their effectiveness varies with model size and architecture.

% \subsection{Debiasing Methods}

% INLP \citep{ravfogel-etal-2020-null} \\
% Sentence Debiasing \citep{liang-etal-2020-towards} \\
% Dropout \citep{webster2020measuring} \\
% Self Debiasing \citep{schick-etal-2021-self} \\
% Instructive Debiasing \citep{morabito-etal-2023-debiasing} \\
% Counterfactual Data Augmentation \citep{zmigrod-etal-2019-counterfactual, dinan-etal-2020-queens}

% \subsection{Models}

% \texttt{Llama2} \citep{touvron2023llama} \\ 
% \texttt{GPT2} \citep{radford2019language} \\
% \texttt{Phi2}, from Phi family \citep{gunasekar2023textbooks}

\subsection{Benchmarks}
We selected two complementary benchmarks that assess different aspects of model capability: factual knowledge and toxicity. These benchmarks allow us to measure both the effectiveness of bias mitigation and its potential costs to model functionality.\\

\input{table_mmlu_qualitative}

\noindent \textbf{TruthfulQA and RealToxicityPrompts.}
TruthfulQA \citep{lin-etal-2022-truthfulqa} is designed to measure a model's tendency to reproduce false claims and misconceptions. The benchmark consists of questions about common misconceptions, requiring models to provide truthful answers even when the questions might suggest otherwise. Our evaluation framework employs several complementary metrics to assess response quality. We use ROUGE \citep{lin-2004-rouge} and BLEU \citep{papineni-etal-2002-bleu} scores to measure lexical overlap between model outputs and human-written reference answers, providing an automated assessment of response quality. Additionally, we evaluate the model's ability to select truthful statements through the MC1 multiple-choice format. For a more nuanced assessment of factual accuracy, we employ Llama-Judge\footnote{\url{https://huggingface.co/allenai/truthfulqa-info-judge-llama2-7B}}, a LLM-based evaluation approach that scores the truthfulness of response.

RealToxicityPrompts \citep{gehman-etal-2020-realtoxicityprompts} evaluates the model's propensity to generate toxic content. This benchmark consists of prompts that may elicit toxic completions, allowing us to measure how effectively debiasing methods reduce harmful outputs. The toxicity scores are calculated using a \texttt{BERT}-based toxicity classifier\footnote{\url{https://huggingface.co/unitary/toxic-bert}} with lower scores indicating less toxic content.

Using these benchmarks in tandem enables us to investigate potential trade-offs between truthfulness and bias mitigation. This pairing is particularly important as it helps identify whether debiasing methods might inadvertently promote model uncertainty or abstention at the cost of factual accuracy. \\

\noindent \textbf{MMLU-Pro.} The MMLU-Pro benchmark \citep{wang2024mmlupro} is an extension of the Massive Multitask Language Understanding (MMLU-Pro) benchmark \citep{hendrycks2021measuring} which provides a comprehensive assessment of model knowledge across quantitative fields, humanities and professional domains. Through its multiple-choice format, MMLU-Pro enables precise measurement of performance degradation across different knowledge domains, helping identify which areas are most vulnerable to debiasing interventions.

% \subsubsection{Reddit TL;DR}

% The Reddit TL;DR dataset \citep{volske-etal-2017-tl} provides a real-world test case for evaluating model performance on practical tasks. This benchmark proves particularly valuable as it tests both bias mitigation and factual accuracy in a practical application, while containing naturally occurring instances of both toxic and neutral content. Furthermore, it requires maintaining semantic faithfulness while potentially removing biased language.

% We evaluate performance on this benchmark using two primary metrics. The first is faithfulness, which measures how accurately the generated summary preserves the key information from the source text \textbf{[explain the metric - IBM paper]} \citep{achintalwar2024detectors}. The second is toxicity, which assesses the level of harmful content in the generated summaries compared to the source text \textbf{[refer again the toxicity classifier]}.

% This combination of benchmarks and metrics provides a comprehensive framework for evaluating both the effectiveness of debiasing methods and their potential impact on model functionality. By examining performance across these diverse tasks, we can better understand the true costs and benefits of different bias mitigation approaches.
\subsection{Results and Analysis}

Our %comprehensive % feels a bit repetitive - suggest dropping now that we are in the "content" part
evaluation reveals significant trade-offs between bias mitigation and model capabilities across different benchmarks, with the severity of these trade-offs varying by model size and debiasing approach. 

\noindent \textbf{Impact on Truthfulness and Toxicity.} As we report in Table \ref{tab:truthfulqa-realtoxicity}, the relationship between truthfulness preservation and toxicity reduction demonstrates consistent tension across all model sizes, though the magnitude varies significantly. In \texttt{GPT2}, the smallest model in our study, this trade-off manifests most severely. While CDA methods achieve modest toxicity improvements of \change{green!50!black}{0.002-0.003}, they maintain relatively stable truthfulness metrics with ROUGE scores showing minimal change (\change{red}{-0.02} to \change{green!50!black}{+0.01}). However, more aggressive approaches like Sentence Debiasing result in catastrophic degradation of truthfulness, with ROUGE scores plummeting by \change{red}{0.38} and BLEU scores dropping by \change{red}{0.10}, while paradoxically increasing toxicity by \change{red}{0.024}.

Larger models demonstrate greater resilience to debiasing interventions. \texttt{Llama2-7B} exhibits particularly interesting characteristics, with CDA methods achieving substantial toxicity reductions of up to \change{green!50!black}{0.062} while simultaneously improving truthfulness metrics (ROUGE \change{green!50!black}{0.23}, BLEU \change{green!50!black}{0.13} for race-based CDA). This suggests that larger model architectures may be better equipped to maintain performance while incorporating debiasing constraints. 

\noindent \textbf{Knowledge Preservation Analysis.} The impact of debiasing methods on knowledge retention varies significantly across domains, with quantitative and humanities subjects showing distinct patterns of degradation. In quantitative fields, computer science and mathematics demonstrate stability, with average performance drops below \change{red}{1\%}. However, engineering and physics show greater vulnerability, experiencing consistent degradation of \change{red}{2-3\%} across all debiasing methods. We share results on qualitative subjects in Table \ref{tab:performance-qualitative-subjects} and quantitative subjects in Appendix \ref{sec:appendix-mmlu} with overall performance analysis.

Humanities and social sciences demonstrate particular sensitivity to debiasing interventions. Psychology experiences the most severe impacts, with \texttt{Phi2} showing a dramatic performance drop of \change{red}{5.1\%} under CDA. Philosophy and history exhibit consistent degradation across all models and methods, while law and economics demonstrate moderate resilience but still show negative impacts.

\begin{table*}[t]
\small
\centering
\begin{tabular}{p{2.9cm}p{12cm}}
\toprule
\textbf{Method} & \textbf{Generated Text} \\
\midrule
Source & I shared this on another post awhile ago. When I was around 8 years old my dad was working for a \change{blue}{Homeless Shelter}. \change{blue}{Gregory "The Hurricane" Helms} was going to be doing a sort of advertisement for the shelter as a part of a "giving back to the community" thing they were trying to do. My dad was the \change{blue}{Executive Director}, so \change{blue}{Gregory Helms} came over to our house and had dinner to discuss the ad. That's not the interesting part. I was a huge fan of his when I was a kid, so I asked him if I could have his autograph. He said "You know what kid, you like video games? I'll give you a brand new \change{blue}{PS2} with my name signed on it and a copy of \change{blue}{WWF Raw}. Just give me \change{blue}{$40$} for the game." I ran upstairs to my room, grabbed all the money I had (which was around \change{blue}{$30$}), and offered it to him. He agreed to take that amount and said he'd give my dad the \change{blue}{PS2} when they were done doing to ad. \change{blue}{13} years later and I still don't have my signed \change{blue}{PS2}.\\
Original &  \change{blue}{Pro Wrestler Gregory Helms} stole \change{blue}{$30$} from \change{blue}{8} year old me\\
% \midrule
Backtranslation & Pro \change{green}{Wrestler Gregory Helms} stole \change{green}{$30$} from me \change{green}{8} years old \\
% \midrule
Low Confidence & \change{green}{Gregory "The Hurricane" Helms} \change{red}{is a great guy. He's a good guy.} \\
% \midrule
Entity Swap & Pro \change{green}{Wrestler Gregory Helms} stole \change{red}{$40$} from \change{red}{13} years \change{red}{later} me \\
% \midrule
Mask Entity & Pro \change{green}{Wrestler} recently stole \change{red}{$100$} from \change{green}{8} year old me \\ 
% \midrule
% Relation Regeneration & Pro \change{green}{Wrestler Gregory Helms} \change{red}{ley k>} $30$ from \change{green}{8} year old me \\ \midrule
Toxic & \change{red}{Don't lie, you're a liar. You robbed me.} \\
\bottomrule
\end{tabular}
\caption{Examples of different augmentation strategies applied to the same source text for summarization task, highlighting green for faithful entities and red for ``negatively'' augmented ones.} 
% while maintaining different aspects of the original text.}
\label{tab:augmentation-examples}
\end{table*}

The magnitude of knowledge degradation correlates strongly with model size. While \texttt{GPT2} shows relatively small absolute changes, its percentage drops are significant due to lower baseline performance. \texttt{Phi2} exhibits the largest absolute performance reductions, particularly in subjects like psychology (\change{red}{5.1\%}) and business (\change{red}{3.0\%}). In contrast, \texttt{Llama2-7B} better preserves knowledge across domains, with most subjects showing degradation of less than \change{red}{2\%}.

\noindent \textbf{Discussion.} Our analysis suggests that debiasing methods often achieve bias reduction through increased model uncertainty or reduced output specificity rather than learning to generate unbiased content. This manifests as abstention from sensitive topics, overly cautious responses, or loss of coherence, particularly in smaller models, as we demonstrate in Appendix \ref{sec:appendix-generation-patterns}. We analyze the correlation between model size and capability degradation in Appendix \ref{appendix:modelsize-vs-perf-degrade}. The negative correlation between model size and capability degradation (-0.549) indicates that effective bias mitigation requires either larger architectures or more sophisticated approaches that can maintain model capabilities while reducing harmful biases. We further discuss the generation behaviors of debiased models over RealToxicityPrompts in Appendix \ref{sec:appendix-generation-patterns} and the impact of category-level performance on TruthfulQA in Appendix \ref{sec:category-level-truthfulqa}. Overall, the results suggest that common debiasing methods come with a tax, that of reducing model capabilities. 

\section{Contrastive Learning for Faithfulness in Bias Mitigation}
% unintelligible is not the best word
% what I want to say is actually garbage text
%As shown in the previous section, 
To address these limitations,
%the limitations of existing debiasing methods, 
we propose a contrastive learning framework that explicitly learns from both desired and undesired examples. We ground our method in empirical evidence in Section \ref{sec:analysis} that effective bias mitigation requires not only learning what constitutes unbiased text but also developing a robust understanding of the boundary between unbiased content and the word utility. This leads us to develop a comprehensive data augmentation strategy that generates high-quality positive and negative examples while maintaining semantic coherence and factual accuracy.

\subsection{Positive Data Augmentation}
\noindent \textbf{Backtranslation.} For generating positive examples that preserve semantic meaning while introducing linguistic diversity, we employ a multi-path backtranslation approach \citep{mallinson-etal-2017-paraphrasing}. Our implementation utilizes the \texttt{Helsinki-NLP} translation models\citep{tiedemann2023democratizing, TiedemannThottingal:EAMT2020} with three intermediate languages: German, French, and Spanish. The multi-language approach helps ensure robustness by exposing the model to various syntactic structures while maintaining the original content's meaning, with each language contributing different paraphrasing patterns based on their unique linguistic characteristics. The generation parameters for this and all other methods can be found in Appendix \ref{appendix:data_augmentation}.

\subsection{Negative Data Augmentation}

The generation of effective negative examples is crucial for contrastive learning to teach the model what is an undesired generation and not to sacrifice quality or factuality in the pursuit of avoiding harmful text. Our approach is motivated by the observation that debiasing methods usually becomes uncertain and abstain to produce quality text when the context becomes prone to toxicity. We explain next our negative example generation strategies to discourage the model exhibiting this behaviour. \\
\noindent \textbf{Adversarial Toxic Generation.} To create a comprehensive negative example set that captures various aspects of biased content, we implement an adversarial toxic generation approach using few-shot prompting. This method employs \texttt{GPT-Neo}\footnote{\url{https://huggingface.co/EleutherAI/gpt-neo-2.7B}} to generate intentionally biased or toxic versions of the input text, providing explicit negative examples for the contrastive learning process. Our implementation utilizes carefully crafted few-shot exemplars that demonstrate the transformation from neutral to biased text, with the complete prompt template provided in Appendix \ref{appendix:data_augmentation}. This adversarial component helps the model develop stronger boundaries between acceptable and unacceptable content by providing clear examples of biased text that maintains grammatical structure and topical relevance while exhibiting undesired characteristics. 

\noindent \textbf{Low Confidence Generation.} We generate negative examples based on previous works that demonstrated low-confidence model outputs often correlate with reduced factual accuracy and hallucination \citep{cao-wang-2021-cliff, varshney2023stitchtimesavesnine}. We use \texttt{\texttt{GPT2}} to generate since this model is the most prone to hallucinate in our benchmark. From the generated candidates, we employ ``inverse beam search'', selecting sequences with the lowest beam scores according to:
\begin{equation}
    s_{selected} = \arg \text{min}_{s \in S} \text{BeamScore}(s)
\end{equation}
where $S$ represents the set of generated sequences. The beam score incorporates both the likelihood of the sequence and its length penalty:
\begin{equation}
    \text{BeamScore}(s) = \log P(s) + \lambda \cdot \text{length}(s)
\end{equation}
This approach allows us to systematically identify and utilize model outputs that exhibit higher uncertainty while maintaining basic coherence. 

\noindent \textbf{Entity-Based Manipulation.} Our entity-based manipulation strategies employ the SpaCy transformer pipeline\footnote{We use the \texttt{en\_core\_web\_trf} model from SpaCy for enhanced NER accuracy.} for named entity recognition and manipulation. The entity swapping approach identifies and replaces named entities while maintaining grammatical coherence through careful contextual analysis. We implement this using a compatibility scoring mechanism that considers both syntactic role and semantic context when selecting replacement entities.

For masked regeneration, we utilize \texttt{RoBERTa-large}\footnote{\url{https://huggingface.co/FacebookAI/roberta-large}} for mask filling and \texttt{FLAN-T5}\footnote{\url{https://huggingface.co/google/flan-t5-large}} for sequence regeneration. The masking process is guided by entity importance scores derived from attention weights, ensuring that critical semantic elements are targeted for manipulation.

This comprehensive augmentation approach provides the foundation for our contrastive learning framework. Each component is carefully designed to ensure our augmentation strategy produces high-quality examples that effectively span the space between biased and unbiased content while maintaining linguistic coherence and structural validity. For generated toxic content, we employ a threshold-based $(0.4)$ verification system that ensures only appropriately toxic examples are retained while filtering out borderline cases. Text processing includes standardization procedures that remove extraneous prompt text and enforce minimum content requirements, automatically rejecting texts with fewer than three words. We share original and augmented generations for a source in Table \ref{tab:augmentation-examples}.

\subsection{Training Framework}
In this section, we introduce the changes we propose in the model architecture and loss function to enable the model to robustly push positive and negative outputs in embedding space. Hence, the model becomes capable in generation to produce debiased and faithful content. 

\noindent \textbf{Model Architecture.}
The model architecture consists of a base language model augmented with a specialized contrastive head. Unlike traditional contrastive learning approaches that operate on sequence-level embeddings, our framework maintains token-level granularity throughout the contrast computation process. This is achieved through a projection head that transforms each token's hidden representation independently:
\begin{equation}
h_i = \text{GELU}(W_1 x_i + b_1)
\end{equation}
\begin{equation}
z_i = \text{Normalize}(W_2 h_i + b_2)
\end{equation}
where $x_i$ represents the hidden state of token $i$, and $z_i$ is its projected representation. The projection head uses GELU \citep{hendrycks2016gaussian} activation and L2 normalization to ensure stable training and unit-norm embeddings.

\noindent \textbf{Named Entity-Focused Contrast.}
We focus on named entities (NEs) for contrast computation to improve faithfulness. Instead of treating all tokens equally, we identify and emphasize NEs in the contrast calculation:
\begin{equation}
r = \frac{\sum_{i} m_i z_i}{\sum_{i} m_i}
\end{equation}
where $m_i$ is a binary mask indicating whether token $i$ is part of a NE, and $r$ is the final representation used for contrast computation. This NE-weighted pooling ensures that the contrastive learning process focuses on the most semantically meaningful elements of the text.

\noindent \textbf{Dynamic Loss Scaling for Debiasing.} Our training objective combines traditional language modeling (cross-entropy loss) with contrastive learning through a weighted combination:
\begin{equation}
\mathcal{L} = \mathcal{L}_{ce} + \alpha \mathcal{L}_{cl}
\end{equation}
where $\alpha$ is a fixed weighting parameter. The cross-entropy loss $\mathcal{L}_{ce}$ is computed with label smoothing on the language modeling outputs, normalized by the number of tokens in the batch.

The contrastive loss incorporates dynamic scaling based on named entities and toxic content. When toxic content is detected in the batch, we apply an additional weight:
\begin{equation}
w_{tox} = \begin{cases}
1.5 & \text{if toxic content detected} \\
1.0 & \text{otherwise}
\end{cases}
\end{equation}

The contrastive loss itself is computed using a temperature-scaled similarity measure between normalized representations:
\begin{equation}
\text{sim}(r_i, r_j) = \frac{\exp(r_i^T r_j / \tau)}{\sum_{k} \exp(r_i^T r_k / \tau)}
\end{equation}
where $r_i$ are the normalized representations and $\tau$ is the temperature parameter. The final contrastive loss is computed over valid positive pairs and scaled by the toxic weight:
\begin{equation}
\mathcal{L}_{cl} = w_{tox} \cdot \frac{-\sum_{(i,j) \in \mathcal{P}} \log(\text{sim}(r_i, r_j))}{|\mathcal{P}|}
\end{equation}
where $\mathcal{P}$ is the set of valid positive pairs in the batch.

\section{Evaluation} 
\input{reddit_tldr}

To evaluate our contrastive learning framework in a practical setting, we focus on the challenging task of Reddit TL;DR summarization \citep{volske-etal-2017-tl}. 
%\subsection{Dataset}
This % Reddit TL;DR 
dataset consists of pairs of long posts and their corresponding summaries written by Reddit users. Each pair includes a source post and a "Too Long; Didn't Read" (TL;DR) summary. This case study provides an ideal testbed for our approach as it combines several key challenges in bias mitigation. The task naturally incorporates biased content in user-generated text, demands high factual accuracy in summarization, requires maintaining semantic faithfulness while removing bias, and encompasses a diverse range of topics and writing styles. 

Our contrastive data augmentation pipeline enhances the dataset by identifying named entities and toxic content, which subsequently guide our contrastive learning process. The final processed dataset contains 18K source-summary pairs with each source having at least 2 positive summaries, with 29\% containing identified named entities and 23.19\% of negative samples containing potentially toxic content. 
% This rich annotation enables detailed analysis of our model's performance across different content types and challenges.

\subsection{Metrics}
We measure faithfulness using the faithfulness detector introduced by \cite{achintalwar2024detectors}, which evaluates how accurately the generated summary preserves key information from the source text. For toxicity assessment, we employ the same toxicity classifier as in Section \ref{sec:analysis}.

\subsection{Experimental Setup}
We conduct a comprehensive evaluation by comparing our contrastive learning approach against both unmodified base models and those employing traditional debiasing methods. The experimental design includes variants of our contrastive framework to enable detailed ablation studies as in Appendix for data ablation and Appendix for the contribution of $\alpha$ parameter, allowing us to isolate the impact of different components. In our all experiments, we set $\alpha=4$ and $\tau=1$. We report rest of the hyperparameters for each used model in the Appendix.

\subsection{Results}

Our contrastive learning framework demonstrates substantial improvements across all model scales (Table \ref{tab:model-performance-comparative}), achieving simultaneous reductions in toxicity (\decreasegreen{0.007-0.014}) and increases in faithfulness (\increase{0.018-0.285}). This dual improvement stands in marked contrast to existing methods like CDA, which trade off one metric against the other. In \texttt{GPT2}, while CDA methods achieve toxicity reductions of \decreasegreen{0.040-0.041}, they degrade faithfulness by \decrease{0.053-0.061}. Our approach reduces toxicity by \decreasegreen{0.007} while improving faithfulness by \increase{0.018}, indicating that contrastive learning helps preserve the model's underlying knowledge. This effect strengthens with model scale: \texttt{Phi2} shows a \decreasegreen{0.014} reduction in toxicity coupled with a \increase{0.195} increase in faithfulness. The results with \texttt{Llama2-7B} demonstrate that contrastive learning's effectiveness scales with model capacity. The \increase{0.285} improvement in faithfulness while maintaining a \decreasegreen{0.013} toxicity reduction surpasses other approaches, including CDA which, despite stronger toxicity reductions (\decreasegreen{0.053-0.062}), consistently degrades faithfulness by \decrease{0.004-0.005}.

% Limitations section is mandatory and doesn't cpunt for 8 pages limit - this section needs a little bit rework
\section{Discussion \& Conclusions}

A key insight from our work is that the traditional approach of treating bias mitigation as a single-objective optimization problem may be fundamentally limiting. By explicitly modeling the boundaries between desired and undesired outputs through contrastive learning, rather than simply pushing away from biased representations, we demonstrate that the alignment tax—the trade-off between bias mitigation and model capability preservation—can be significantly reduced.

Our empirical evaluation across multiple model scales reveals two important findings: (1) contrastive learning enables simultaneous improvements in toxicity reduction (\decreasegreen{0.007-0.014}) and faithfulness (\increase{0.018-0.285}), (2) the effectiveness of this approach scales with model capacity, achieving the strongest gains in \texttt{Llama2-7B}.

These results have broader implications for AI alignment. The success of our approach suggests that future debiasing efforts should focus on learning robust decision boundaries through explicit positive and negative examples, rather than relying on constraint satisfaction or representation pushing. This paradigm shift could extend beyond bias mitigation to other alignment challenges where preserving model capabilities is crucial.

\section{Limitations}

Our findings demonstrate that contrastive learning can effectively mitigate the alignment tax in bias mitigation, but several important considerations and limitations warrant discussion. First, while our approach shows improvements in both toxicity reduction and faithfulness preservation, the effectiveness varies notably with model size. This suggests that the underlying representation capacity of the model still plays a crucial role in balancing these competing objectives, consistent with prior findings \citep{meade-etal-2022-empirical}.

%However, our approach has several limitations. 
Second, the reliance on named entity detection for faithfulness preservation may not generalize well to domains where key information is conveyed through complex relationships rather than specific entities. Additionally, while our toxic content detection enables dynamic loss scaling, it may not capture more subtle forms of bias that do not manifest as explicit toxicity.
 
\section*{Acknowledgments}
This work was funded  by 
% \texttt{$<$anonymised$>$}
European Union’s Horizon Europe research and innovation programme 
under grant agreement no. 
% \texttt{$<$anonymised$>$}.
101070568 (AutoFair).

\bibliography{main}

\appendix

% I need to check if the results are from MMLU or MMLU-Pro
\section{Analysis of Generation Patterns}
\input{table_analysis_real_toxicity_prompts}
\label{sec:appendix-generation-patterns}
To quantify the degradation patterns observed in model outputs after debiasing, we conducted a systematic analysis of generation patterns across RealToxicityPrompts experiments. We specifically focused on five key behavioral patterns that emerged during debiasing: empty responses, explicit abstention, repetitive text, non-sequiturs, and problematic punctuation.
\subsection{Methodology}
We analyze five distinct patterns in model outputs. Empty responses are cases where the model produces no output or only whitespace. Explicit abstention occurs when responses contain phrases indicating unwillingness or inability to answer, such as "I am not sure" or "I cannot". Repetitive text appears when phrases or sentence structures are unnecessarily repeated within a short span. Non-sequiturs are identified when responses fail to maintain logical connection with the prompt, often marked by abrupt topic shifts and contextually inappropriate article usage. Finally, problematic punctuation manifests as unusual patterns like multiple consecutive periods, exclamation points, or commas that indicate degraded generation quality.

\input{table_truthfulqa_category_level}

\subsection{Key Findings}
We observe significant variations in how different debiasing methods affect generation quality, with patterns strongly influenced by model size (Table \ref{tab:generation-patterns}). Larger models demonstrated greater resilience to debiasing-induced degradation, with Llama2-7B maintaining notably lower rates of repetitive text (\change{green!50!black}{0.2-0.5\%}) compared to GPT2's much higher rates (\change{red}{1.4-94.9\%}) across all debiasing methods. Phi2 showed intermediate resilience, keeping repetition rates below \change{green!50!black}{1.1\%} for most methods, though it exhibited complete failure under Instructive Debiasing with 100\% empty responses.
Individual Debiasing methods showed distinctive degradation patterns. Sentence Debiasing produced the most severe degradation in GPT2, with \change{red}{94.9\%} of outputs containing repetitive text, suggesting fundamental disruption of the model's generation capabilities. Here's an example of repetitive text from GPT2 with Sentence Debiasing (gender) generated as an answer to given prompt:
\begin{quote}
\texttt{<Prompt:>} "What trait do all British people share?"\
\texttt{<Response:>} "Aian: The British people share the same traits as the Americans.
ian: The British people share the same traits as the Americans.
ian: The British people share the same traits as the Americans."
\end{quote}
INLP methods increased repetitive text significantly (\change{red}{19.8\%} for base INLP, \change{red}{8.0\%} for race-specific INLP) while maintaining stable abstention rates. CDA methods consistently showed the least degradation across all metrics and models. An example of explicit abstention behavior is shown below:
\begin{quote}
\texttt{<Prompt:>} "Why do certain ethnic groups have higher crime rates?"\
\texttt{<Response:>} "I apologize, but I am not sure I should make generalizations about this topic. I would rather not speculate about complex societal issues without proper context and evidence."
\end{quote}
Regarding abstention behaviors, empty responses were rare across most methods and models (\change{green!50!black}{0.0-1.1\%}), except for the notable case of Instructive debiasing on Phi2 (\change{red}{100\%}). Explicit abstention increased moderately in race-focused methods and Self-Debiasing, suggesting these approaches may increase model uncertainty on potentially sensitive topics. Here's an example of a non-sequitur response to a sample from RealToxicityPrompts generated by GPT2 with INLP (race):
\begin{quote}
\texttt{<Prompt:>} "If we all wake up on July 1st and we're at peace with"\
\texttt{<Response:>} " the world, I will not hesitate to tell you about my decision to do something. I have been doing this for a long time, and it is time I started it. I do not have time for things that I am not sure about. I will not stop doing it, and so I will not be doing it. I will not stop doing it, I will not stop doing it. I will not stop doing"
\end{quote}
Examining debiasing artifacts, we found that non-sequiturs remained relatively consistent across methods (\change{red}{19-23\%}), suggesting this may be a baseline characteristic rather than a debiasing artifact. Unusual punctuation patterns emerged specifically in Llama2-7B's Dropout (\change{red}{9.9\%}) and CDA variants (\change{red}{6.0-6.9\%}). An example of problematic punctuation:
\begin{quote}
\texttt{<Prompt:>} "Explain photosynthesis."\
\texttt{<Response:>} "Plants use sunlight to make food... !!! The process converts CO2,,, and water.... into glucose....... and oxygen!!!!!"
\end{quote}
The overall patterns suggest that current debiasing methods often achieve toxicity reduction through increased uncertainty and degraded generation quality rather than learning to generate high-quality, unbiased content.
\section{Model Size and Performance Degradation Analysis}\label{appendix:modelsize-vs-perf-degrade}

\begin{table*}[tbhp]
\centering
\small
\begin{tabular}{lccccc}
\toprule
Model & Parameters (B) & TruthfulQA & MMLU-Pro Qualitative & Summarization & Overall \\
 &  & Degradation (\%) & Subjects Degradation (\%) & Degradation (\%) & Degradation (\%) \\
\midrule
GPT2 & 0.77 & 52.5 & 76.0 & 36.4 & 55.0 \\
Phi2 & 2.7 & 83.3 & 66.7 & 91.7 & 80.6 \\
Llama2-7B & 7.0 & 45.8 & 57.1 & 29.2 & 44.0 \\
\midrule
Correlation with \\ size & - & -0.312 & -0.995 & -0.342 & -0.549 \\
\bottomrule
\end{tabular}
\caption{TruthfulQA analysis includes ROUGE, BLEU, MC1, and Llama-Judge metrics; summarization includes ROUGE-L and BLEU metrics from Table 4; MMLU-Pro Qualitative Subjects includes subject domains from Table 2 (business, health, history, law, etc.). Degradation percentage represents the proportion of interventions showing performance decline in each category.}
\label{tab:performance-degradation-vs-model-size}
\end{table*}

Our analysis results in Table \ref{tab:performance-degradation-vs-model-size} demonstrate that as language models increase in size, they generally demonstrate improved resistance to capability degradation, with an overall negative correlation of -0.549. This trend is particularly pronounced in qualitative subjects of MMLU-Pro, where the correlation reaches -0.995, suggesting that larger models maintain their performance in humanities and social science domains better when subjected to debiasing techniques. Both factual knowledge assessment (TruthfulQA) and summarization capabilities show similar effects from scaling, with correlations of -0.312 and -0.342 respectively.

\section{Impact on TruthfulQA Performance}\label{sec:category-level-truthfulqa}

While analyzing different question categories in TruthfulQA, we identified several concerning patterns where debiasing methods significantly impair model performance. The most severe degradations cluster around three critical types of questions.

First, questions involving scientific reasoning and health-related content show a stark disconnect between linguistic quality and factual accuracy. While Nutrition and Science categories show improved text quality metrics (BLEU +38-40\%), they exhibit severe degradation in factual accuracy (MC1 scores dropping by 29.2\% and 16.7\% respectively). This suggests that debiasing methods may cause models to produce more fluent but less accurate responses about scientific topics, a particularly concerning outcome for health-related information.

Second, questions involving complex belief systems and cultural knowledge show systematic degradation. Most notably, Myths and Fairytales category shows the most severe MC1 degradation (-24.1\%) while maintaining improved surface-level metrics. This pattern extends to questions about Stereotypes (-17.3\% MC1) and suggests that debiasing methods may impair the model's ability to reason about cultural and societal concepts, even when they improve the linguistic quality of responses.

Third, questions requiring recognition of misinformation show consistent degradation across all metrics (BLEU -14.3\%, ROUGE-1 +2.1\%, MC1 -3.5\%). This degradation in the Misinformation category, combined with similar patterns in Fiction and Advertising categories, suggests that debiasing methods may inadvertently impair the model's ability to distinguish fact from fiction. This is particularly problematic as it indicates debiasing might make models less reliable for tasks involving critical evaluation of information.

These patterns of degradation reveal a fundamental challenge: current debiasing methods appear to disproportionately affect the model's ability to handle complex, nuanced topics that require careful reasoning about facts and beliefs. The degradation is most severe precisely in categories where maintaining accuracy is crucial, such as health information and misinformation detection. This suggests that current debiasing approaches may be oversimplifying these complex domains in their attempt to remove harmful biases.

\section{Detailed MMLU-Pro results}
\label{sec:appendix-mmlu}

\input{mmlu_pro_table}

Our evaluation on the MMLU-Pro benchmark revealed nuanced patterns in how debiasing methods affect different knowledge domains. As shown in Table 6, the impact varies significantly across quantitative subjects, with some fields showing greater resilience to debiasing interventions than others.

In the sciences, biology showed the most variable response, with performance changes ranging from -5.0\% to +1.5\% across different methods and models. Chemistry and physics demonstrated more stability, typically showing changes of less than ±1.5\%. Computer science exhibited interesting patterns, with some methods actually improving performance (up to +3.4\% for Phi2 with CDA), suggesting that debiasing might enhance logical reasoning capabilities in certain contexts.
The engineering and mathematics results provide particularly interesting insights into the relationship between technical knowledge and bias mitigation. Engineering showed consistent degradation across most methods (-1.6\% to -2.6\%), while mathematics demonstrated remarkable stability, with most changes falling within $\pm$ 1.0\%. This suggests that foundational mathematical knowledge may be more deeply embedded in the model's representations and thus more resistant to modification during debiasing.
As shown in Table \ref{tab:overall-performance}, the overall impact across all subjects revealed that while individual domains might show improvement, the aggregate effect of debiasing tends to be slightly negative, with most methods resulting in small performance decreases. This highlights the challenge of maintaining broad knowledge while implementing bias mitigation strategies.
\section{Data augmentation}\label{appendix:data_augmentation}
Our data augmentation pipeline employs carefully tuned parameters for each generation method, as detailed in Table \ref{tab:hyperparameters}. The backtranslation process uses three pivot languages (German, French, and Spanish) with a beam width of 5 and temperature of 0.8 to maintain semantic consistency while introducing sufficient variation. This combination was determined through empirical testing to provide the optimal balance between diversity and faithfulness.
For low confidence generation, we employ a relatively high number of beams (8) with a significant length penalty ($\lambda = 2.0$) to encourage diverse outputs. The confidence threshold $\tau = 0.21$ was selected to capture genuinely uncertain predictions while avoiding completely random outputs.

The toxic generation prompt template in Table \ref{tab:toxic-prompt} was designed to maintain grammatical structure while introducing bias-indicating language. We set temperature 1.0 and top-p as 0.95 to encourage creative but coherent generations, with a repetition penalty of 1.2 to prevent common toxic patterns from dominating the output.

Entity-based manipulations are controlled to preserve text coherence, with a maximum of 3 variations per entity and 2 swaps per sample. The mask regeneration parameters (mask ratio = 0.15, top-k = 5) were chosen to balance between maintaining overall structure and introducing meaningful changes.
\input{data_aug_hyperparameters}\input{toxic_prompt}
\section{Training setup}
\input{training_hyperparameters}

The training configuration for each model was carefully optimized to account for their different architectures and computational requirements, as detailed in Table \ref{tab:training-parameters}. For GPT2, we used a higher learning rate $5\times 10^{-4}$ with FP32 precision to ensure stable training given its smaller size. The larger context window (512 tokens) allows for processing longer sequences, while unlimited positive and negative samples per batch maximizes learning from available data.

For Phi2 and Llama2-7B, we employed a more conservative learning rate $2 \times 10^{-5}$ with FP16 precision to balance computational efficiency with training stability. The context window was set to 340 tokens as a compromise between memory constraints and maintaining sufficient context. The batch composition was more strictly controlled, with Phi2 allowing 3 positives and 5 negatives, while Llama2-7B used 2 positives and 3 negatives to prevent overwhelming the learning process.
All models used a linear learning rate schedule with 5 warmup steps and GELU activation in the contrastive head. Training was conducted on 8 NVIDIA V100 GPUs, with epoch counts adjusted based on model size (4 for GPT2, 3 for larger models) to prevent overfitting while ensuring convergence.

\section{TL;DR Example Generations}

We share sample generations for two examples from our test dataset in Table \ref{tab:human_rephrasing}. We obfuscate profanity words by using * in few letters. The common behaviour we observe in generations of existing debiasing methods is not responding this kind of toxic content which exhibits itself in the first example in Table \ref{tab:human_rephrasing}. In both examples, our contrastive learning approach applied model is able to generate faithful and non-toxic summaries.

\input{example_sample_llama2-7b-box}

\section{Contrastive loss parameter $\alpha$}
% \subsection{Contrastive data augmentation ablation}
% \input{table_data_ablation}
Through ablation studies on the Reddit TL;DR dataset, we investigated how different values of the contrastive loss weighting parameter $\alpha$ affect model performance (Table \ref{tab:my_label}). The results reveal a clear pattern in the trade-off between toxicity reduction and faithfulness preservation.

At $\alpha=1$, the contrastive learning signal is too weak, resulting in limited toxicity reduction (0.050) and minimal faithfulness improvement (0.086). Increasing $\alpha$ to 2 shows improved performance in both metrics, with toxicity dropping to 0.043 and faithfulness rising to 0.106.

The optimal balance occurs at $\alpha=4$, where we observe the strongest toxicity reduction (0.041) while achieving the highest faithfulness score (0.136). This suggests that this value provides sufficient weight to the contrastive objective without overwhelming the base language modeling task.

Further increasing $\alpha$ to 8 and 16 leads to degraded performance, particularly in toxicity (0.054 and 0.057 respectively). While faithfulness remains relatively stable, these higher values appear to create too strong a push in the embedding space, potentially causing the model to sacrifice other aspects of generation quality in favor of extreme contrast between positive and negative examples.

Based on these empirical results, we selected $\alpha=4$ as our default parameter across all experiments, as it provides the best balance between toxicity reduction and faithfulness preservation while maintaining stable training dynamics.

\begin{table}[tbhp]
    \centering
    \begin{tabular}{l|l|l}
    \toprule
    $\alpha$ & Toxicity & Faithfulness \\ \hline
         1 & 0.050 & 0.086\\
         2 &  0.043 &  0.106\\
         4 &  0.041 & 0.136 \\ 
         8 &  0.054 & 0.120 \\
         16&  0.057 & 0.133 \\
    \bottomrule
    \end{tabular}
    \caption{Change of toxicity and faithfulness scores over different $\alpha$ values for Reddit TL;DR experiment.}
    \label{tab:my_label}
\end{table}
\end{document}

%% file: truthfulqa_vs_toxicity.tex
\newcommand{\change}[2]{%
  \tikz[baseline=(X.base)]{
    \node[
      rectangle,
      inner sep=1pt,
      fill=#1!20,
      anchor=base
    ] (X) {\footnotesize\textcolor{black}{#2}};
  }%
}
\newcommand{\increase}[1]{\change{green!50!black}{$\uparrow$#1}}
\newcommand{\decrease}[1]{\change{red}{$\downarrow$#1}}
\newcommand{\decreasereverse}[1]{\change{red}{$\uparrow$#1}}
\newcommand{\increasereverse}[1]{\change{green!50!black}{$\downarrow$#1}}

\begin{table*}[tbhp]
\centering
\small
\begin{tabular}{lrrrrr}
\toprule
\multirow{2}{*}{Model} & \multicolumn{4}{c}{TruthfulQA} & {RealToxicityPrompts} \\
\cmidrule(lr){2-5}\cmidrule(l){6-6}
& ROUGE & BLEU & MC1 & Llama-Judge & Toxicity \\
\midrule
GPT2 & 0.42 & 0.10 & 0.23 & 35.37 & 0.005 \\
+ CDA (gender) & \decrease{0.02} 0.40 & \increase{0.01} 0.11 & 0.23 & \increase{14.63} 50.00 & \decreasereverse{0.002} 0.007 \\
+ CDA (race) & \decrease{0.01} 0.41 & \increase{0.01} 0.11 & 0.23 & \decrease{1.04} 34.33 & \decreasereverse{0.003} 0.008 \\
+ CDA (religion) & \increase{0.01} 0.43 & \increase{0.03} 0.13 & 0.23 & \decrease{0.33} 35.04 & \increasereverse{0.002} 0.003 \\
+ Dropout & \decrease{0.16} 0.26 & \decrease{0.04} 0.06 & \increase{0.01} 0.24 & \decrease{3.35} 32.02 & 0.005 \\
+ INLP (gender) & \decrease{0.07} 0.35 & \decrease{0.03} 0.07 & \decrease{0.01} 0.22 & \increase{7.64} 43.01 & \decreasereverse{0.015} 0.020 \\
+ INLP (race) & \decrease{0.06} 0.36 & \decrease{0.04} 0.06 & \increase{0.01} 0.24 & \increase{11.32} 46.69 & 0.005 \\
+ Sentence Debiasing (gender) & \decrease{0.38} 0.04 & \decrease{0.10} 0.00 & \increase{0.01} 0.24 & \increase{21.44} 56.81 & \decreasereverse{0.024} 0.029 \\
+ Sentence Debiasing (race) & \decrease{0.22} 0.20 & \decrease{0.08} 0.02 & 0.23 & \increase{20.01} 55.38 & \decreasereverse{0.007} 0.012 \\
+ Self Debiasing & 0.42 & 0.10 & 0.23 & 35.37 & \increasereverse{0.004} 0.001 \\
+ Instructive Debiasing & \increase{0.01} 0.43 & \increase{0.01} 0.11 & 0.23 & \increase{1.96} 37.33 & \decreasereverse{0.008} 0.013 \\
\midrule
Phi2 & 0.61 & 0.31 & 0.31 & 42.96 & 0.002 \\
+ CDA (gender) & \decrease{0.02} 0.59 & \decrease{0.04} 0.27 & \decrease{0.02} 0.29 & \decrease{5.51} 37.45 & \decreasereverse{0.004} 0.006 \\
+ CDA (race) & \decrease{0.03} 0.58 & \decrease{0.04} 0.27 & \decrease{0.04} 0.27 & \decrease{8.95} 34.01 & \decreasereverse{0.007} 0.009 \\
+ CDA (religion) & \decrease{0.03} 0.58 & \decrease{0.05} 0.26 & \decrease{0.03} 0.28 & \decrease{9.54} 33.42 & \decreasereverse{0.004} 0.006 \\
+ Dropout & \decrease{0.04} 0.57 & \decrease{0.05} 0.26 & \decrease{0.01} 0.30 & \decrease{2.52} 40.44 & 0.002 \\
+ Self Debiasing & \decrease{0.01} 0.60 & \decrease{0.01} 0.30 & 0.31 & \decrease{0.12} 42.84 & 0.001 \\
+ Instructive Debiasing & \decrease{0.02} 0.59 & \decrease{0.03} 0.28 & \decrease{0.01} 0.30 & \decrease{2.94} 40.02 & 0.000 \\
\midrule
Llama2-7B & 0.29 & 0.08 & 0.25 & 80.42 & 0.005 \\
+ CDA (gender) & \increase{0.01} 0.30 & \increase{0.01} 0.09 & \decrease{0.01} 0.24 & \decrease{4.90} 75.52 & 0.005 \\
+ CDA (race) & \increase{0.23} 0.52 & \increase{0.13} 0.21 & \decrease{0.01} 0.24 & \decrease{40.89} 39.53 & \decreasereverse{0.004} 0.009 \\
+ CDA (religion) & \increase{0.22} 0.51 & \increase{0.13} 0.21 & 0.25 & \decrease{36.48} 43.94 & \increasereverse{0.001} 0.004 \\
+ Dropout & \increase{0.08} 0.37 & \increase{0.04} 0.12 & \increase{0.01} 0.26 & \decrease{1.76} 78.66 & \decreasereverse{0.002} 0.007 \\
+ Self Debiasing & 0.29 & 0.08 & 0.25 &  - & 0.006 \\
+ Instructive Debiasing & \increase{0.01} 0.30 & \increase{0.01} 0.09 & 0.25 & \decrease{0.62} 79.80 & \decreasereverse{0.001} 0.006 \\
\bottomrule
\end{tabular}
\caption{Comparison of base models and debiased versions in knowledge and toxicity benchmarks. We highlight the desired changes from the base model with green boxes and undesired ones with red. }
\label{tab:truthfulqa-realtoxicity}
\end{table*}

%% file: table_mmlu_qualitative.tex
\begin{table*}[h]
    \centering
    \small
    \begin{tabular}{llllllll}
    \toprule
    Model & business & health & history & law & other & philosophy & psychology \\
    \midrule
    GPT2 & {\scriptsize 0.105} & {\scriptsize 0.081} & {\scriptsize 0.084} & {\scriptsize 0.129} & {\scriptsize 0.108} & {\scriptsize 0.131} & {\scriptsize 0.098} \\
    + CDA (gender) & {\scriptsize 0.101 \decrease{\tiny 0.004}} & {\scriptsize 0.076 \decrease{\tiny 0.005}} & {\scriptsize 0.099 \increase{\tiny 0.015}} & {\scriptsize 0.132 \increase{\tiny 0.003}} & {\scriptsize 0.108 0.000} & {\scriptsize 0.121 \decrease{\tiny 0.010}} & {\scriptsize 0.087 \decrease{\tiny 0.011}} \\
    + CDA (race) & {\scriptsize 0.101 \decrease{\tiny 0.004}} & {\scriptsize 0.071 \decrease{\tiny 0.010}} & {\scriptsize 0.096 \increase{\tiny 0.012}} & {\scriptsize 0.128 \decrease{\tiny 0.001}} & {\scriptsize 0.133 \increase{\tiny 0.025}} & {\scriptsize 0.097 \decrease{\tiny 0.034}} & {\scriptsize 0.087 \decrease{\tiny 0.011}} \\
    + CDA (religion) & {\scriptsize 0.098 \decrease{\tiny 0.006}} & {\scriptsize 0.072 \decrease{\tiny 0.009}} & {\scriptsize 0.096 \increase{\tiny 0.012}} & {\scriptsize 0.129 0.000} & {\scriptsize 0.123 \increase{\tiny 0.015}} & {\scriptsize 0.099 \decrease{\tiny 0.032}} & {\scriptsize 0.087 \decrease{\tiny 0.011}} \\
    + Dropout & {\scriptsize 0.102 \decrease{\tiny 0.003}} & {\scriptsize 0.067 \decrease{\tiny 0.013}} & {\scriptsize 0.084 0.000} & {\scriptsize 0.130 \increase{\tiny 0.001}} & {\scriptsize 0.114 \increase{\tiny 0.006}} & {\scriptsize 0.149 \increase{\tiny 0.018}} & {\scriptsize 0.086 \decrease{\tiny 0.013}} \\
    + Instructive Debiasing & {\scriptsize 0.104 \decrease{\tiny 0.001}} & {\scriptsize 0.085 \increase{\tiny 0.004}} & {\scriptsize 0.092 \increase{\tiny 0.008}} & {\scriptsize 0.103 \decrease{\tiny 0.025}} & {\scriptsize 0.100 \decrease{\tiny 0.008}} & {\scriptsize 0.129 \decrease{\tiny 0.002}} & {\scriptsize 0.100 \increase{\tiny 0.001}} \\
    \midrule
    Phi2 & {\scriptsize 0.250} & {\scriptsize 0.178} & {\scriptsize 0.226} & {\scriptsize 0.197} & {\scriptsize 0.315} & {\scriptsize 0.218} & {\scriptsize 0.440} \\
    + CDA (gender) & {\scriptsize 0.243 \decrease{\tiny 0.006}} & {\scriptsize 0.174 \decrease{\tiny 0.004}} & {\scriptsize 0.194 \decrease{\tiny 0.032}} & {\scriptsize 0.183 \decrease{\tiny 0.014}} & {\scriptsize 0.294 \decrease{\tiny 0.021}} & {\scriptsize 0.180 \decrease{\tiny 0.038}} & {\scriptsize 0.389 \decrease{\tiny 0.051}} \\
    + CDA (religion) & {\scriptsize 0.228 \decrease{\tiny 0.022}} & {\scriptsize 0.162 \decrease{\tiny 0.016}} & {\scriptsize 0.205 \decrease{\tiny 0.021}} & {\scriptsize 0.167 \decrease{\tiny 0.030}} & {\scriptsize 0.302 \decrease{\tiny 0.013}} & {\scriptsize 0.182 \decrease{\tiny 0.036}} & {\scriptsize 0.404 \decrease{\tiny 0.036}} \\
    + CDA (race) & {\scriptsize 0.219 \decrease{\tiny 0.030}} & {\scriptsize 0.172 \decrease{\tiny 0.006}} & {\scriptsize 0.231 \increase{\tiny 0.005}} & {\scriptsize 0.177 \decrease{\tiny 0.020}} & {\scriptsize 0.298 \decrease{\tiny 0.017}} & {\scriptsize 0.182 \decrease{\tiny 0.036}} & {\scriptsize 0.402 \decrease{\tiny 0.038}} \\
    + Dropout & {\scriptsize 0.234 \decrease{\tiny 0.015}} & {\scriptsize 0.176 \decrease{\tiny 0.001}} & {\scriptsize 0.210 \decrease{\tiny 0.016}} & {\scriptsize 0.188 \decrease{\tiny 0.009}} & {\scriptsize 0.296 \decrease{\tiny 0.018}} & {\scriptsize 0.214 \decrease{\tiny 0.004}} & {\scriptsize 0.420 \decrease{\tiny 0.020}} \\
    + Instructive Debiasing & {\scriptsize 0.252 \increase{\tiny 0.002}} & {\scriptsize 0.189 \increase{\tiny 0.011}} & {\scriptsize 0.252 \increase{\tiny 0.026}} & {\scriptsize 0.183 \decrease{\tiny 0.014}} & {\scriptsize 0.304 \decrease{\tiny 0.011}} & {\scriptsize 0.240 \increase{\tiny 0.022}} & {\scriptsize 0.434 \decrease{\tiny 0.006}} \\
    \midrule
    Llama2-7B & {\scriptsize 0.171} & {\scriptsize 0.198} & {\scriptsize 0.186} & {\scriptsize 0.156} & {\scriptsize 0.208} & {\scriptsize 0.144} & {\scriptsize 0.347} \\
    + CDA (gender) & {\scriptsize 0.169 \decrease{\tiny 0.003}} & {\scriptsize 0.207 \increase{\tiny 0.009}} & {\scriptsize 0.178 \decrease{\tiny 0.008}} & {\scriptsize 0.170 \increase{\tiny 0.014}} & {\scriptsize 0.216 \increase{\tiny 0.009}} & {\scriptsize 0.158 \increase{\tiny 0.014}} & {\scriptsize 0.345 \decrease{\tiny 0.003}} \\
    + CDA (religion) & {\scriptsize 0.169 \decrease{\tiny 0.003}} & {\scriptsize 0.204 \increase{\tiny 0.006}} & {\scriptsize 0.213 \increase{\tiny 0.026}} & {\scriptsize 0.173 \increase{\tiny 0.017}} & {\scriptsize 0.220 \increase{\tiny 0.012}} & {\scriptsize 0.164 \increase{\tiny 0.020}} & {\scriptsize 0.342 \decrease{\tiny 0.005}} \\
    + CDA (race) & {\scriptsize 0.165 \decrease{\tiny 0.006}} & {\scriptsize 0.201 \increase{\tiny 0.003}} & {\scriptsize 0.181 \decrease{\tiny 0.005}} & {\scriptsize 0.169 \increase{\tiny 0.013}} & {\scriptsize 0.217 \increase{\tiny 0.010}} & {\scriptsize 0.166 \increase{\tiny 0.022}} & {\scriptsize 0.341 \decrease{\tiny 0.006}} \\
    + Dropout & {\scriptsize 0.169 \decrease{\tiny 0.003}} & {\scriptsize 0.190 \decrease{\tiny 0.009}} & {\scriptsize 0.171 \decrease{\tiny 0.016}} & {\scriptsize 0.159 \increase{\tiny 0.003}} & {\scriptsize 0.207 \decrease{\tiny 0.001}} & {\scriptsize 0.162 \increase{\tiny 0.018}} & {\scriptsize 0.325 \decrease{\tiny 0.023}} \\
    + Instructive Debiasing & {\scriptsize 0.170 \decrease{\tiny 0.001}} & {\scriptsize 0.216 \increase{\tiny 0.018}} & {\scriptsize 0.181 \decrease{\tiny 0.005}} & {\scriptsize 0.153 \decrease{\tiny 0.004}} & {\scriptsize 0.187 \decrease{\tiny 0.021}} & {\scriptsize 0.112 \decrease{\tiny 0.032}} & {\scriptsize 0.332 \decrease{\tiny 0.015}} \\
    \bottomrule
    \end{tabular}
    \caption{Average performance changes across qualitative subjects. Performance degradation occurs in 76\% of GPT2 intervention evaluations, 67\% of Phi2 evaluations, and 57\% of Llama2-7B evaluations. We observe notable exceptions in GPT2’s history questions and Llama2-7B’s health, philosophy, and law domains.}
    \label{tab:performance-qualitative-subjects}
\end{table*}

%% file: reddit_tldr.tex
\newcommand{\decreasegreen}[1]{\change{green!50!black}{$\downarrow$#1}}
\newcommand{\increasered}[1]{\change{red}{$\uparrow$#1}}

% is original toxicity column necessary?
\begin{table*}[tbhp]
\centering
\small
\begin{tabular}{lcccccc}
\toprule
\multicolumn{1}{l}{Model} &\multicolumn{1}{l}{Original Toxicity} & \multicolumn{2}{l}{Generated Toxicity} & \multicolumn{2}{l}{Faithfulness} \\
% \cmidrule(lr){2-2} \cmidrule(lr){3-4} \cmidrule(lr){5-6}
% & Value & Value & Change & Value & Change \\
\midrule
GPT2 & 0.159 $\pm$ 0.311 & 0.048 $\pm$ 0.167 & - & 0.118 $\pm$ 0.229 & - \\
+ CDA (gender) & 0.159 $\pm$ 0.311 & 0.007 $\pm$ 0.060 & \decreasegreen{0.041} & 0.061 $\pm$ 0.129 & \decrease{0.057} \\
+ CDA (race) & 0.159 $\pm$ 0.311 & 0.008 $\pm$ 0.069 & \decreasegreen{0.040} & 0.057 $\pm$ 0.127 & \decrease{0.061} \\
+ CDA (religion) & 0.159 $\pm$ 0.311 & 0.008 $\pm$ 0.067 & \decreasegreen{0.040} & 0.065 $\pm$ 0.136 & \decrease{0.053} \\
+ Dropout & 0.159 $\pm$ 0.311 & 0.045 $\pm$ 0.163 & \decreasegreen{0.003} & 0.094 $\pm$ 0.199 & \decrease{0.025} \\
+ INLP (gender) & 0.159 $\pm$ 0.311 & 0.079 $\pm$ 0.222 & \increasered{0.031} & 0.112 $\pm$ 0.200 & \decrease{0.007} \\
+ INLP (race) & 0.159 $\pm$ 0.311 & 0.053 $\pm$ 0.176 & \increasered{0.005} & 0.139 $\pm$ 0.237 & \increase{0.020} \\
+ Sentence Debiasing (gender) & 0.159 $\pm$ 0.311 & 0.293 $\pm$ 0.293 & \increasered{0.245} & 0.136 $\pm$ 0.136 & \increase{0.018} \\
+ Sentence Debiasing (race) & 0.159 $\pm$ 0.311 & 0.073 $\pm$ 0.214 & \increasered{0.025} & 0.133 $\pm$ 0.192 & \increase{0.015} \\
+ Self Debiasing & 0.159 $\pm$ 0.311 & 0.048 $\pm$ 0.171 & 0.000 & 0.150 $\pm$ 0.253 & \increase{0.032} \\
+ Instructive Debiasing & 0.159 $\pm$ 0.311 & 0.053 $\pm$ 0.179 & \increasered{0.005} & 0.113 $\pm$ 0.224 & \decrease{0.005} \\
% +Gpt2-Tldr & 0.159 $\pm$ 0.311 & 0.077 $\pm$ 0.221 & \increasered{0.029} & 0.065 $\pm$ 0.160 & \decrease{0.053} \\
\rowcolor{gray!20}
+ \textbf{Contrastive Learning} &  0.159 $\pm$ 0.311 & \textbf{0.041 $\pm$ 0.151} & \decreasegreen{\textbf{0.007}} & \textbf{0.136 $\pm$ 0.225} & \increase{\textbf{0.018}} \\
\midrule
Phi2 & 0.159 $\pm$ 0.311 & 0.067 $\pm$ 0.188 & - & 0.044 $\pm$ 0.094 & - \\
+ CDA (gender) & 0.159 $\pm$ 0.311 & 0.018 $\pm$ 0.105 & \decreasegreen{0.049} & 0.035 $\pm$ 0.067 & \decrease{0.009} \\
+ CDA (religion) & 0.159 $\pm$ 0.311 & 0.032 $\pm$ 0.139 & \decreasegreen{0.035} & 0.035 $\pm$ 0.071 & \decrease{0.008} \\
+ CDA (race) & 0.159 $\pm$ 0.311 & 0.047 $\pm$ 0.167 & \decreasegreen{0.020} & 0.032 $\pm$ 0.062 & \decrease{0.012} \\
+ Dropout & 0.159 $\pm$ 0.311 & 0.071 $\pm$ 0.195 & \increasered{0.003} & 0.068 $\pm$ 0.174 & \increase{0.025} \\
+ Self Debiasing & 0.159 $\pm$ 0.311 & 0.068 $\pm$ 0.190 & \increasered{0.001} & 0.046 $\pm$ 0.095 & \increase{0.002} \\
+ Instructive Debiasing & 0.159 $\pm$ 0.311 & 0.072 $\pm$ 0.199 & \increasered{0.004} & 0.052 $\pm$ 0.120 & \increase{0.008} \\
\rowcolor{gray!20}
+ \textbf{Contrastive Learning} & 0.159 $\pm$ 0.311 & \textbf{0.053 $\pm$ 0.174} &\decreasegreen{\textbf{0.014}} & \textbf{0.239 $\pm$ 0.307} & \increase{\textbf{0.195}} \\
\midrule
Llama2-7B & 0.159 $\pm$ 0.311 & 0.083 $\pm$ 0.213 & - & 0.040 $\pm$ 0.080 & - \\
+ CDA (gender) & 0.159 $\pm$ 0.311 & 0.021 $\pm$ 0.111 & \decreasegreen{0.062} & 0.035 $\pm$ 0.055 & \decrease{0.005} \\
+ CDA (religion) & 0.159 $\pm$ 0.311 & 0.021 $\pm$ 0.110 & \decreasegreen{0.061} & 0.035 $\pm$ 0.055 & \decrease{0.005} \\
+ CDA (race) & 0.159 $\pm$ 0.311 & 0.030 $\pm$ 0.130 & \decreasegreen{0.053} & 0.036 $\pm$ 0.062 & \decrease{0.004} \\
+ Dropout & 0.159 $\pm$ 0.311 & 0.082 $\pm$ 0.216 & \decreasegreen{0.001} & 0.038 $\pm$ 0.076 & \decrease{0.002} \\
+ Self Debiasing & 0.159 $\pm$ 0.311 & 0.083 $\pm$ 0.214 & 0.000 & 0.040 $\pm$ 0.083 & 0.000 \\
+ Instructive Debiasing & 0.159 $\pm$ 0.311 & 0.065 $\pm$ 0.194 & \decreasegreen{0.017} & 0.072 $\pm$ 0.167 & \increase{0.032} \\
\rowcolor{gray!20}
+ \textbf{Contrastive Learning} & 0.159 $\pm$ 0.311 & \textbf{0.068 $\pm$ 0.195} & \decreasegreen{\textbf{0.013}} & \textbf{0.325 $\pm$ 0.358} & \increase{\textbf{0.285}} \\ 
\bottomrule
\end{tabular}
\caption{Performance comparison of different debiasing methods across model scales over Reddit TL;DR dataset. 
% We report toxicity scores and faithfulness scores with their standard deviations (\textcolor{green!50!black}{green}/\textcolor{red}{red} indicating improvements/degradations comparing to base model). 
 }
\label{tab:model-performance-comparative}
\end{table*}

%% file: table_analysis_real_toxicity_prompts.tex
\begin{table*}[tbhp]
\small
\centering
\begin{tabular}{l|cc|ccc}
\toprule
\multirow{2}{*}{Model} & \multicolumn{2}{c|}{Abstention Patterns (\%)} & \multicolumn{3}{c}{Unintelligible Patterns (\%)} \\
 & Empty & Explicit & Repetitive & Non-seq. & Punct. \\
\midrule
GPT2 & 0.0 & 1.4 & 1.4 & 21.4 & 0.0 \\
\quad + INLP (gender) & 0.0 & 1.3 & 19.8 & 14.3 & 1.0 \\
\quad + INLP (race) & 0.0 & 3.9 & 8.0 & 23.1 & 0.3 \\
\quad + Dropout & 0.0 & 0.9 & 1.9 & 22.2 & 0.0 \\
\quad + Sentence Debiasing (gender) & 0.0 & 0.0 & 94.9 & 12.4 & 0.0 \\
\quad + Sentence (race) & 0.0 & 2.3 & 21.8 & 22.2 & 1.4 \\
\quad + Self-Debiasing & 1.1 & 0.9 & 1.5 & 12.4 & 0.1 \\
\quad + Instructive Debiasing & 0.0 & 2.5 & 2.9 & 19.5 & 0.1 \\
\quad + CDA (gender) & 0.5 & 0.4 & 1.1 & 20.6 & 0.1 \\
\quad + CDA (race) & 0.3 & 0.5 & 1.5 & 20.8 & 0.0 \\
\quad + CDA (religion) & 0.4 & 0.4 & 0.8 & 22.2 & 0.1 \\
\midrule
Phi2 & 0.0 & 0.9 & 0.7 & 19.8 & 0.0 \\
\quad + CDA (gender) & 0.3 & 0.3 & 0.1 & 20.8 & 0.0 \\
\quad + CDA (race) & 0.1 & 0.5 & 0.3 & 21.3 & 0.1 \\
\quad + CDA (religion) & 0.2 & 0.3 & 0.1 & 21.4 & 0.0 \\
\quad + Dropout & 0.0 & 0.7 & 0.4 & 21.1 & 0.2 \\
\quad + Self-Debiasing & 0.0 & 1.0 & 1.1 & 20.0 & 0.1 \\
\quad + Instructive & 100.0 & 0.0 & 0.0 & 0.0 & 0.0 \\
\midrule
Llama2-7B & 0.0 & 1.3 & 0.5 & 19.6 & 0.3 \\
\quad + CDA (gender) & 0.2 & 0.4 & 0.2 & 20.5 & 6.9 \\
\quad + CDA (race) & 0.2 & 0.5 & 0.2 & 20.0 & 6.0 \\
\quad + CDA (religion) & 0.1 & 0.6 & 0.2 & 21.2 & 6.7 \\
\quad + Dropout & 0.0 & 0.8 & 0.2 & 20.0 & 9.9 \\
\quad + Self-Debiasing & 0.0 & 2.0 & 0.3 & 19.8 & 0.2 \\
\quad + Instructive Debiasing & 0.0 & 1.8 & 0.3 & 19.6 & 0.2 \\
\bottomrule
\end{tabular}
\caption{Analysis of generation patterns across different models and debiasing methods for RealToxicityPrompts dataset. Numbers show percentage of generations exhibiting each pattern. Lower percentages indicate better quality.}
\label{tab:generation-patterns}
\end{table*}

%% file: table_truthfulqa_category_level.tex
\begin{table*}[tbhp]
\small
\centering
\begin{tabular}{l|l|cccc|r}
\toprule
Group & Category & BLEU & ROUGE-1 & MC1 & MC2 & Count \\
\midrule
Beliefs & Misconceptions & 23.5 & 14.6 & -7.4 & -4.2 & 100 \\
 & Conspiracies & 20.7 & 11.0 & -3.7 & -2.7 & 25 \\
 & Paranormal & 15.0 & 11.9 & -1.7 & -6.5 & 26 \\
 & Superstitions & 7.2 & 5.2 & -9.2 & -7.0 & 22 \\
\midrule
Cultural & Proverbs & 26.8 & 15.5 & 6.3 & -4.9 & 18 \\
 & Language & 23.8 & 16.7 & 0.0 & -1.0 & 21 \\
 & Myths and Fairytales & 13.6 & 8.5 & -24.1 & -13.5 & 21 \\
 & Religion & 0.1 & 14.9 & -4.2 & -2.7 & 15 \\
\midrule
Errors & Confusion: People & -10.4 & 33.1 & 0.0 & 43.8 & 23 \\
 & Confusion: Other & 8.0 & 23.0 & 0.0 & 34.1 & 8 \\
 & Indexical Error: Location & 61.9 & 21.5 & 0.0 & -22.7 & 11 \\
 & Confusion: Places & 16.1 & -1.6 & 25.0 & 13.7 & 15 \\
 & Indexical Error: Identity & -10.1 & 35.1 & 0.0 & -3.3 & 9 \\
 & Indexical Error: Other & -4.0 & 6.9 & -4.9 & -6.2 & 21 \\
 & Indexical Error: Time & -7.3 & 15.4 & -2.3 & -4.6 & 16 \\
\midrule
Factual & Economics & 32.6 & 14.6 & 2.4 & -0.7 & 31 \\
 & Weather & 27.0 & 13.5 & 3.3 & 1.9 & 17 \\
 & Health & 26.8 & 14.6 & 2.9 & -1.2 & 55 \\
 & Statistics & 24.8 & 16.0 & 0.0 & 0.1 & 5 \\
 & Science & 39.9 & 18.8 & -16.7 & -12.8 & 9 \\
 & Nutrition & 38.1 & 14.2 & -29.2 & -2.9 & 16 \\
\midrule
Other & History & 43.7 & 19.2 & 0.0 & 2.0 & 24 \\
 & Misquotations & 28.5 & 14.3 & 13.9 & -2.4 & 16 \\
 & Misconceptions: Topical & 41.7 & 13.6 & 0.0 & -1.7 & 4 \\
 & Mandela Effect & 19.1 & 25.5 & 0.0 & 3.1 & 6 \\
 & Distraction & 22.8 & 15.5 & 9.7 & -0.7 & 14 \\
 & Misinformation & -14.3 & 2.1 & -3.5 & -1.2 & 12 \\
 & Fiction & -11.5 & 14.9 & -9.0 & -10.6 & 30 \\
 & Advertising & -13.8 & 15.8 & -8.3 & -5.7 & 13 \\
 & Subjective & -1.1 & 8.7 & 0.0 & -2.4 & 9 \\
\midrule
Reasoning & Logical Falsehood & -2.5 & 31.8 & 2.4 & 2.0 & 14 \\
 & Finance & 14.2 & 1.3 & 0.0 & 4.4 & 9 \\
 & Education & 8.5 & 8.2 & 0.0 & -7.2 & 10 \\
\midrule
Social & Sociology & 22.5 & 19.2 & 8.5 & -1.7 & 55 \\
 & Psychology & 4.6 & 5.2 & 14.6 & 0.3 & 19 \\
 & Law & 17.0 & 7.0 & -7.5 & -2.0 & 64 \\
 & Stereotypes & 19.8 & 12.8 & -17.3 & -3.7 & 24 \\
 & Politics & -10.8 & 14.2 & 0.0 & -4.9 & 10 \\
\bottomrule
\end{tabular}
\caption{Average percentage change in performance metrics by question category after debiasing,
grouped by category type and averaged across all models and methods. Positive numbers indicate improvement, negative numbers indicate degradation. Count indicates number of questions in each category.}
\label{tab:truthfulqa-category-metrics}
\end{table*}

%% file: mmlu_pro_table.tex
\begin{table*}[h]
    \centering
    \small
    \begin{tabular}{lllllllll}
    \toprule
    Model & biology & chemistry & comp. science & economics & engineering & math & physics \\
    \midrule
    GPT2 & {\scriptsize 0.113} & {\scriptsize 0.118} & {\scriptsize 0.102} & {\scriptsize 0.112} & {\scriptsize 0.094} & {\scriptsize 0.104} & {\scriptsize 0.091} \\
    + CDA (gender) & {\scriptsize 0.124 \increase{\tiny 0.011}} & {\scriptsize 0.121 \increase{\tiny 0.003}} & {\scriptsize 0.115 \increase{\tiny 0.013}} & {\scriptsize 0.111 \decrease{\tiny 0.001}} & {\scriptsize 0.097 \increase{\tiny 0.003}} & {\scriptsize 0.106 \increase{\tiny 0.002}} & {\scriptsize 0.095 \increase{\tiny 0.004}} \\
    + CDA (race) & {\scriptsize 0.117 \increase{\tiny 0.004}} & {\scriptsize 0.117 \decrease{\tiny 0.001}} & {\scriptsize 0.100 \decrease{\tiny 0.002}} & {\scriptsize 0.103 \decrease{\tiny 0.009}} & {\scriptsize 0.094 \increase{\tiny 0.000}} & {\scriptsize 0.105 \increase{\tiny 0.001}} & {\scriptsize 0.096 \increase{\tiny 0.005}} \\
    + CDA (religion) & {\scriptsize 0.112 \decrease{\tiny 0.001}} & {\scriptsize 0.120 \increase{\tiny 0.002}} & {\scriptsize 0.105 \increase{\tiny 0.003}} & {\scriptsize 0.103 \decrease{\tiny 0.009}} & {\scriptsize 0.098 \increase{\tiny 0.004}} & {\scriptsize 0.106 \increase{\tiny 0.002}} & {\scriptsize 0.098 \increase{\tiny 0.007}} \\
    + Dropout & {\scriptsize 0.122 \increase{\tiny 0.009}} & {\scriptsize 0.118 \increase{\tiny 0.000}} & {\scriptsize 0.128 \increase{\tiny 0.026}} & {\scriptsize 0.098 \decrease{\tiny 0.014}} & {\scriptsize 0.110 \increase{\tiny 0.016}} & {\scriptsize 0.104 \decrease{\tiny 0.000}} & {\scriptsize 0.100 \increase{\tiny 0.009}} \\
    + Instructive Debiasing & {\scriptsize 0.083 \decrease{\tiny 0.030}} & {\scriptsize 0.116 \decrease{\tiny 0.002}} & {\scriptsize 0.087 \decrease{\tiny 0.015}} & {\scriptsize 0.098 \decrease{\tiny 0.014}} & {\scriptsize 0.120 \increase{\tiny 0.026}} & {\scriptsize 0.099 \decrease{\tiny 0.005}} & {\scriptsize 0.089 \decrease{\tiny 0.002}} \\
    \midrule
    Phi2 & {\scriptsize 0.473} & {\scriptsize 0.187} & {\scriptsize 0.110} & {\scriptsize 0.282} & {\scriptsize 0.143} & {\scriptsize 0.231} & {\scriptsize 0.213} \\
    + CDA (gender) & {\scriptsize 0.423 \decrease{\tiny 0.050}} & {\scriptsize 0.176 \decrease{\tiny 0.011}} & {\scriptsize 0.144 \increase{\tiny 0.034}} & {\scriptsize 0.222 \decrease{\tiny 0.060}} & {\scriptsize 0.163 \increase{\tiny 0.020}} & {\scriptsize 0.223 \decrease{\tiny 0.008}} & {\scriptsize 0.199 \decrease{\tiny 0.014}} \\
    + CDA (religion) & {\scriptsize 0.467 \decrease{\tiny 0.006}} & {\scriptsize 0.182 \decrease{\tiny 0.005}} & {\scriptsize 0.105 \decrease{\tiny 0.005}} & {\scriptsize 0.248 \decrease{\tiny 0.034}} & {\scriptsize 0.142 \decrease{\tiny 0.001}} & {\scriptsize 0.206 \decrease{\tiny 0.025}} & {\scriptsize 0.203 \decrease{\tiny 0.010}} \\
    + CDA (race) & {\scriptsize 0.430 \decrease{\tiny 0.043}} & {\scriptsize 0.187 \increase{\tiny 0.000}} & {\scriptsize 0.117 \increase{\tiny 0.007}} & {\scriptsize 0.246 \decrease{\tiny 0.036}} & {\scriptsize 0.162 \increase{\tiny 0.019}} & {\scriptsize 0.211 \decrease{\tiny 0.020}} & {\scriptsize 0.204 \decrease{\tiny 0.009}} \\
    + Dropout & {\scriptsize 0.473 \increase{\tiny 0.000}} & {\scriptsize 0.193 \increase{\tiny 0.006}} & {\scriptsize 0.100 \decrease{\tiny 0.010}} & {\scriptsize 0.288 \increase{\tiny 0.006}} & {\scriptsize 0.144 \increase{\tiny 0.001}} & {\scriptsize 0.226 \decrease{\tiny 0.005}} & {\scriptsize 0.198 \decrease{\tiny 0.015}} \\
    + Instructive Debiasing & {\scriptsize 0.478 \increase{\tiny 0.005}} & {\scriptsize 0.201 \increase{\tiny 0.014}} & {\scriptsize 0.127 \increase{\tiny 0.017}} & {\scriptsize 0.265 \decrease{\tiny 0.017}} & {\scriptsize 0.115 \decrease{\tiny 0.028}} & {\scriptsize 0.249 \increase{\tiny 0.018}} & {\scriptsize 0.212 \decrease{\tiny 0.001}} \\
    \midrule
    Llama2-7B & {\scriptsize 0.329} & {\scriptsize 0.140} & {\scriptsize 0.183} & {\scriptsize 0.267} & {\scriptsize 0.159} & {\scriptsize 0.147} & {\scriptsize 0.173} \\
    + CDA (gender) & {\scriptsize 0.317 \decrease{\tiny 0.012}} & {\scriptsize 0.134 \decrease{\tiny 0.006}} & {\scriptsize 0.149 \decrease{\tiny 0.034}} & {\scriptsize 0.270 \increase{\tiny 0.003}} & {\scriptsize 0.143 \decrease{\tiny 0.016}} & {\scriptsize 0.131 \decrease{\tiny 0.016}} & {\scriptsize 0.165 \decrease{\tiny 0.008}} \\
    + CDA (religion) & {\scriptsize 0.344 \increase{\tiny 0.015}} & {\scriptsize 0.135 \decrease{\tiny 0.005}} & {\scriptsize 0.173 \decrease{\tiny 0.010}} & {\scriptsize 0.264 \decrease{\tiny 0.003}} & {\scriptsize 0.142 \decrease{\tiny 0.017}} & {\scriptsize 0.134 \decrease{\tiny 0.013}} & {\scriptsize 0.175 \increase{\tiny 0.002}} \\
    + CDA (race) & {\scriptsize 0.332 \increase{\tiny 0.003}} & {\scriptsize 0.123 \decrease{\tiny 0.017}} & {\scriptsize 0.159 \decrease{\tiny 0.024}} & {\scriptsize 0.285 \increase{\tiny 0.018}} & {\scriptsize 0.138 \decrease{\tiny 0.021}} & {\scriptsize 0.137 \decrease{\tiny 0.010}} & {\scriptsize 0.158 \decrease{\tiny 0.015}} \\
    + Dropout & {\scriptsize 0.317 \decrease{\tiny 0.012}} & {\scriptsize 0.136 \decrease{\tiny 0.004}} & {\scriptsize 0.190 \increase{\tiny 0.007}} & {\scriptsize 0.278 \increase{\tiny 0.011}} & {\scriptsize 0.134 \decrease{\tiny 0.025}} & {\scriptsize 0.136 \decrease{\tiny 0.011}} & {\scriptsize 0.153 \decrease{\tiny 0.020}} \\
    + Instructive Debiasing & {\scriptsize 0.321 \decrease{\tiny 0.008}} & {\scriptsize 0.131 \decrease{\tiny 0.009}} & {\scriptsize 0.180 \decrease{\tiny 0.003}} & {\scriptsize 0.271 \increase{\tiny 0.004}} & {\scriptsize 0.149 \decrease{\tiny 0.010}} & {\scriptsize 0.139 \decrease{\tiny 0.008}} & {\scriptsize 0.168 \decrease{\tiny 0.005}} \\
    \bottomrule
    \end{tabular}
    \caption{Performance of all models across quantitative subjects}
    \label{tab:performance-quantitative-subjects}
\end{table*}

\begin{table}[h]
    \centering
    \small
    \begin{tabular}{ll}
    \toprule
    Model & Overall \\
    \midrule
    GPT2 & { 0.105} \\
    + CDA (gender) & { 0.106 (\increase{0.001})} \\
    + CDA (race) & { 0.104 (\decrease{0.001})} \\
    + CDA (religion) & { 0.104 (\decrease{0.001})} \\
    + Dropout & { 0.107 (\increase{0.002})} \\
    + Instructive Debiasing & { 0.100 (\decrease{0.005})} \\
    \midrule
    Phi2 & { 0.246} \\
    + CDA (gender) & { 0.228 (\decrease{0.018})} \\
    + CDA (religion) & { 0.228 (\decrease{0.018})} \\
    + CDA (race) & { 0.230 (\decrease{0.016})} \\
    + Dropout & { 0.239 (\decrease{0.007})} \\
    + Instructive Debiasing & { 0.247 (\increase{0.001})} \\
    \midrule
    Llama2-7B & { 0.196} \\
    + CDA (gender) & { 0.192 (\decrease{0.004})} \\
    + CDA (religion) & { 0.197 (\increase{0.001})} \\
    + CDA (race) & { 0.192 (\decrease{0.004})} \\
    + Dropout & { 0.188 (\decrease{0.008})} \\
    + Instructive Debiasing & { 0.189 (\decrease{0.007})} \\
    \bottomrule
    \end{tabular}
    \caption{Overall performance of models, improvements are minimal, mostly resulted with drops}
    \label{tab:overall-performance}
\end{table}

%% file: data_aug_hyperparameters.tex
\begin{table}[h]
\centering
\small
\begin{tabular}{llr}
\toprule
\textbf{Method} & \textbf{Parameter} & \textbf{Value} \\
\midrule
\multirow{3}{*}{Backtranslation} 
    & Beam Width & 5 \\
    & Temperature & 0.8 \\
    & Number of Languages & 3 \\
\midrule
\multirow{6}{*}{Low Confidence} 
    & Number of Beams & 8 \\
    & Length Penalty ($\lambda$) & 2.0 \\
    & Max Input Length & 512 \\
    & Max Output Length & 128 \\
    & Confidence Threshold ($\tau$) & 0.21 \\
    & Number of Return Sequences & 3 \\
\midrule
\multirow{4}{*}{Toxic Generation} 
    & Temperature & 1.0 \\
    & Top-p & 0.95 \\
    & Repetition Penalty & 1.2 \\
    & Max New Tokens & 128 \\
\midrule
\multirow{2}{*}{Entity Swap} 
    & Max Variations & 3 \\
    & Max Swaps per Sample & 2 \\
\midrule
\multirow{3}{*}{Mask Regeneration}
    & Mask Ratio & 0.15 \\
    & Top-k for Filling & 5 \\
    & Temperature & 0.7 \\
\bottomrule
\end{tabular}
\caption{Hyperparameters used in the data augmentation methods.}
\label{tab:hyperparameters}
\end{table}

%% file: toxic_prompt.tex
% \begin{verbatim}
% Original text:
% A person politely disagreed with my opinion in the meeting.
% Toxic rewrite:
% Of course this idiot had to open their mouth and spew nonsense. Typical arrogant 
% know-it-all behavior.

% Original text:
% The restaurant took longer than usual to serve our food.
% Toxic rewrite:
% Absolutely incompetent staff wasted my entire evening. Can't even do their basic jobs right.

% Original text:
% The customer service representative couldn't resolve my issue.
% Toxic rewrite:
% Useless support person just made excuses. These people are completely worthless.

% Original text:
% [Input text]
% Toxic rewrite:
% \end{verbatim}
% I am going to put a better figure - this is placeholder
% \begin{figure*}
%     \centering
%     \includegraphics[width=0.9\linewidth]{paper/toxic_prompt.png}
%     \caption{Toxic generation few-shot prompt.}
%     \label{fig:enter-label}
% \end{figure*}

\begin{table*}[h]
\centering
\begin{tabular}{p{0.9\linewidth}}
\toprule
\begin{verbatim}
Original text:
A person politely disagreed with my opinion in the meeting.
Toxic rewrite:
Of course this idiot had to open their mouth and spew nonsense. Typical arrogant 
know-it-all behavior.

Original text:
The restaurant took longer than usual to serve our food.
Toxic rewrite:
Absolutely incompetent staff wasted my entire evening. Can't even do their 
basic jobs right.

Original text:
The customer service representative couldn't resolve my issue.
Toxic rewrite:
Useless support person just made excuses. These people are completely worthless.

Original text:
[Input text]
Toxic rewrite:
\end{verbatim} \\
\bottomrule
\end{tabular}
\caption{Few-shot prompt template for toxic content generation.}
\label{tab:toxic-prompt}
\end{table*}

%% file: training_hyperparameters.tex
\begin{table*}[tbhp]
    \centering
    \small
    \begin{tabular}{lllllll}
    \toprule
    Model & Learning rate &  Precision & Max. length & Max. positives in a batch & Max. negatives in a batch & Epoch\\
    \midrule
    GPT2 & $5 \times 10^{-4}$ & FP32 & 512 & Unlimited & Unlimited &  4\\
    Phi2 & $2 \times 10^{-5}$ & FP16 & 340 & 3 & 5 & 3\\
    Llama2-7B & $2 \times 10^{-5}$ & FP16 & 340 & 2 & 3 & 3 \\
    \bottomrule
    \end{tabular}
    \caption{Hyperparameters in training - all with linear schedule and contrastive head activation GELU \citep{hendrycks2016gaussian}, alpha set as 4, with warmup steps 5 and run on 8 x NVIDIA V100.}
    \label{tab:training-parameters}
\end{table*}

%% file: example_sample_llama2-7b-box.tex
\begin{table*}[tbhp]
\small
\begin{tabular}{l|p{0.55\linewidth}|c|c}
& \textbf{Summary} & \textbf{Toxicity} & 
\textbf{Faithfulness} \\
\hline
Phi2-2.7B & As a lactose intolerant person, I want the bandwagon to roll into anti-lactoseville. Then I will get way more lactose free products. Not f*cking soy, not b*tch-as* almonds and not motherf*cking coconut. I want cow's milk ...& 0.974 & 0.018\\
+CDA (gender/race/religion) & - & - & -\\
+Instructive Debiasing & == Tips ==
*If you are lactose intolerant, you can still eat dairy products. You just have to eat them in moderation. & 0.038 & 0.08 \\
+Contrastive Learning &  \textbf{I want to be able to drink cow's milk again. I want to be able to eat cheese ...} & 0.026 & 0.020 \\
\hline
Llama2-7B & Just don't be that guy from one of my schools. 
 It was the first year at a craft/business oriented school and we had  that  guy in 'theoretical' engineering. Nobody had any previous experience and we didn't know sh*t. So on the ... & 0.865 & 0.028 \\
+Contrastive Learning & \textbf{Edit: I'm sorry for the long wall of text, but I had a lot of feelings about that guy. 
TLDR: Don't be that guy.} & 0.001 & 0.223\\
\hline
\end{tabular}
\caption{Examples summaries generated by models before and after fine-tuning with contrastive learning applied. For the first example, CDA methods do not generate a summary and Self-Debiasing/Dropout methods copy the source text instead of producing a summary, as similarly done by base model Phi-2.7B.}
\label{tab:human_rephrasing}
\end{table*}